\documentclass[10pt,twocolumn,letterpaper]{article}

\usepackage[pagenumbers]{cvpr} % To force page numbers, e.g. for an arXiv version

\usepackage{graphicx}
\usepackage{amsmath}
\usepackage{amssymb}
\usepackage{booktabs}
\usepackage{bm}
\usepackage{epsfig}
\makeatletter
\@namedef{ver@everyshi.sty}{}
\makeatother
\usepackage{tikz}
\usetikzlibrary{spy}
\usepackage{algpseudocode}
\usepackage{algorithm}
\usepackage{mathrsfs}
\usepackage{nicefrac}       % compact symbols for 1/2, etc.
\usepackage{booktabs}       % professional-quality tables
\usepackage{adjustbox}
\usepackage{makecell}
\usepackage{tabularx}
\usepackage{tikz, pgfplots}
\usetikzlibrary{positioning}

\usepackage[pagebackref,breaklinks,colorlinks]{hyperref}

\newcommand{\bb}{{\boldsymbol b}}

\newcommand{\xb}{{\boldsymbol x}}
\newcommand{\yb}{{\boldsymbol y}}
\newcommand{\zb}{{\boldsymbol z}}
\newcommand{\wb}{{\boldsymbol w}}
\newcommand{\nb}{{\boldsymbol n}}
\newcommand{\fb}{{\boldsymbol f}}

\newcommand{\Ab}{{\boldsymbol A}}
\newcommand{\Db}{{\boldsymbol D}}
\newcommand{\Ib}{{\boldsymbol I}}

\newcommand{\Ed}{{\mathbb E}}
\newcommand{\Rd}{{\mathbb R}}

\newcommand{\Nc}{{\mathcal N}}
\newcommand{\Sc}{{\mathcal S}}
\newcommand{\Pc}{{\mathcal P}}

\DeclareMathOperator*{\argmin}{\arg\!\min}

\newcommand{\code}[1] {\texttt{#1}}

\newcommand{\tickYtopD}[1]{\raisebox{-1.5ex}[0ex][0ex]{#1}}
\newcommand{\tickPSNR}{\tickYtopD{PSNR[db]}}
\newcommand{\tickNview}[1]{\smash{\# view$=$}{$#1$}\hspace*{1em}}

\definecolor{C0}{rgb}{0.121569, 0.466667, 0.705882}
\definecolor{C1}{rgb}{1.000000, 0.498039, 0.054902}
\definecolor{C2}{rgb}{0.172549, 0.627451, 0.172549}
\definecolor{C3}{rgb}{0.839216, 0.152941, 0.156863}
\definecolor{C4}{rgb}{0.580392, 0.403922, 0.741176}
\definecolor{C5}{rgb}{0.549020, 0.337255, 0.294118}
\definecolor{C6}{rgb}{0.890196, 0.466667, 0.760784}
\definecolor{C7}{rgb}{0.498039, 0.498039, 0.498039}
\definecolor{C8}{rgb}{0.737255, 0.741176, 0.133333}
\definecolor{C9}{rgb}{0.090196, 0.745098, 0.811765}
\definecolor{trolleygrey}{rgb}{0.5, 0.5, 0.5}

\usepackage[capitalize]{cleveref}
\crefname{section}{Sec.}{Secs.}
\Crefname{section}{Section}{Sections}
\Crefname{table}{Table}{Tables}
\crefname{table}{Tab.}{Tabs.}

\def\cvprPaperID{7272} % *** Enter the CVPR Paper ID here
\def\confName{CVPR}
\def\confYear{2023}

\newcommand{\s}{{\boldsymbol s}}

\begin{document}

\title{Solving 3D Inverse Problems using Pre-trained 2D Diffusion Models}

\author{Hyungjin Chung\textsuperscript{\rm 1,2}\thanks{Equal contribution}{ \ }, Dohoon Ryu\textsuperscript{\rm 1}$^{*}$,  Michael T. Mccann\textsuperscript{\rm 2}, Marc L. Klasky\textsuperscript{\rm 2}, Jong Chul Ye\textsuperscript{\rm 1}\\
\textsuperscript{\rm 1}Korea Advanced Institute of Science \& Technology, 
\textsuperscript{\rm 2}Los Alamos National Laboratory,\\
{\tt\small \{hj.chung, dh.ryu, jong.ye\}@kaist.ac.kr, \{mccann, mklasky\}@lanl.gov}}
% cover figure
\twocolumn[{%
\renewcommand\twocolumn[1][]{#1}%
\maketitle
\begin{center}
\vspace{-1.0cm}
% \includegraphics[width=\linewidth]{./figs/cover.jpg}
% \captionof{figure}{(a) Algorithmic overview of DiffusionMBIR: parallel denoising followed by prior-augmenting ADMM-TV, (b) representative results: 4-view sparse-view CT reconstruction.}
\includegraphics[width=\linewidth]{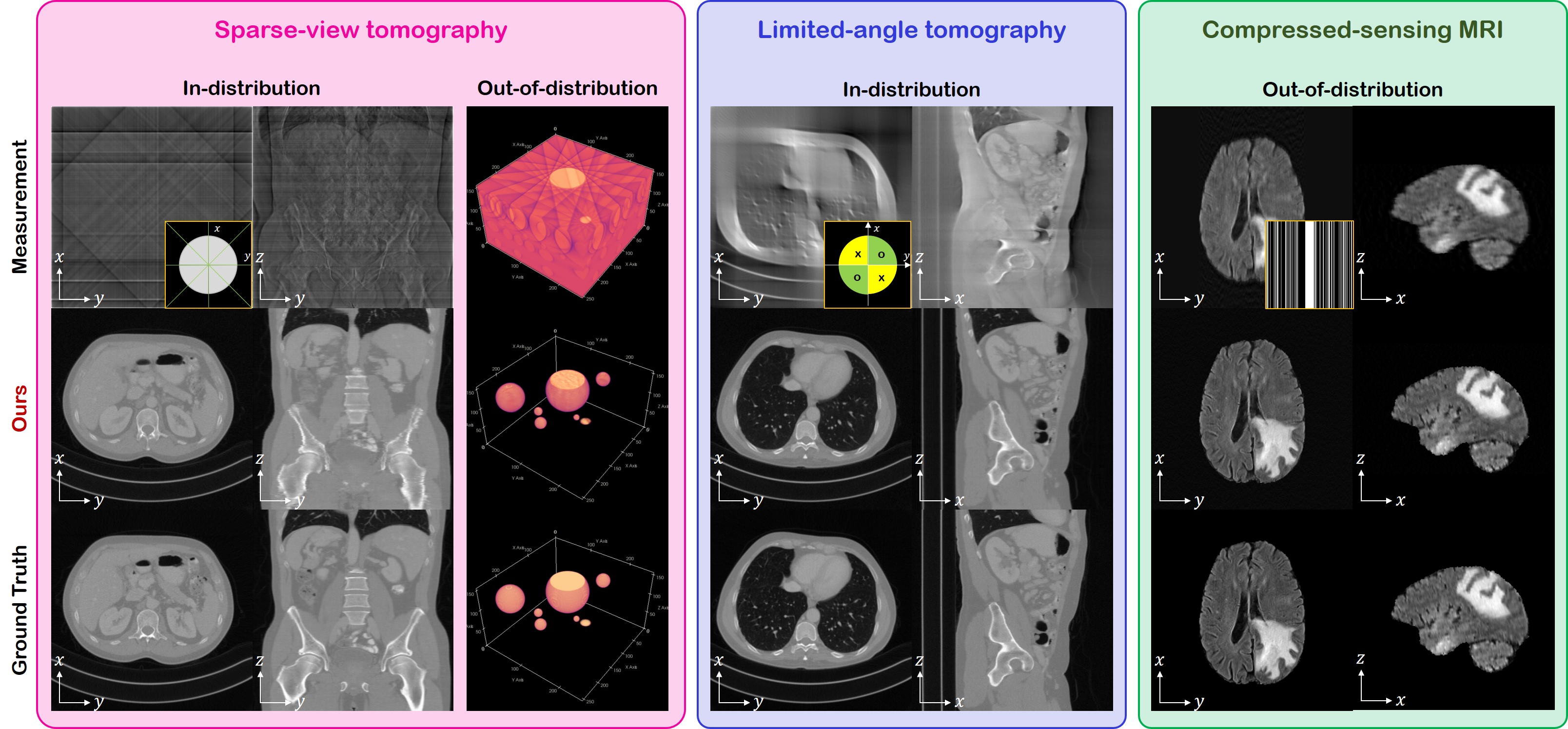}
\captionof{figure}{3D reconstruction results with DiffusionMBIR. First row: measurement, second row: our method, third row: ground truth. Yellow inset: measurement process. Sparse-view tomography: 8-view measurement, Limited-angle tomography: [0 90]$^\circ$ out of [0 180]$^\circ$ angle measurement, Compressed-sensing MRI: 1D uniform sub-sampling of $\times 2$ acceleration. (In-distribution): test data aligned with training data, (Out-of-distribution): test data vastly different from training data.}
\label{fig:cover}
\end{center}
}]

% \maketitle

%%%%%%%%% ABSTRACT
\begin{abstract}
\vspace{-0.5cm}
Diffusion models have emerged as the new state-of-the-art generative model with high quality samples, with intriguing properties such as mode coverage and high flexibility. They have also been shown to be effective inverse problem solvers, acting as the prior of the distribution, while the information of the forward model can be granted at the sampling stage. Nonetheless, as the generative process remains in the same high dimensional (i.e. identical to data dimension) space, the models have not been extended to 3D inverse problems due to the extremely high memory and computational cost. In this paper, we combine the ideas from the conventional model-based iterative reconstruction with the modern diffusion models, which leads to a highly effective method for solving 3D medical image reconstruction tasks such as sparse-view tomography, limited angle tomography, compressed sensing MRI from pre-trained 2D diffusion models. In essence, we propose to augment the 2D diffusion prior with a model-based prior in the remaining direction at test time, such that one can achieve coherent reconstructions across all dimensions.
Our method can be run in a single commodity GPU, and establishes the new state-of-the-art, showing that the proposed method can perform reconstructions of high fidelity and accuracy even in the most extreme cases (e.g. 2-view 3D tomography). 
We further reveal that the generalization capacity of the proposed method is surprisingly high, and can be used to reconstruct volumes that are entirely different from the training dataset.
\vspace{-0.8cm}
\end{abstract}

\begin{figure*}[!t]
    \centering
    \includegraphics[width=1.0\textwidth]{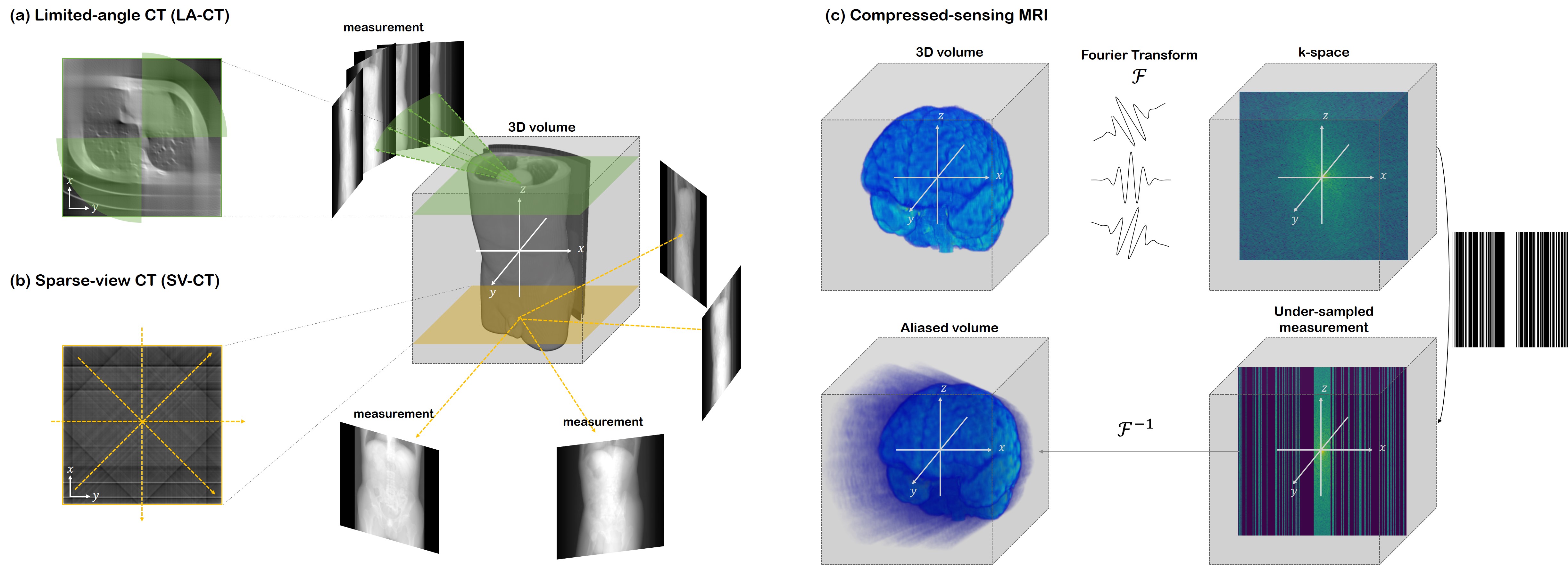}
    \caption{Visualization of the measurement process for the three tasks we tackle in this work: (a) Limited angle CT (LA-CT)---measurement model of Fig.~\ref{fig:lact_results}, (b) sparse view CT (SV-CT)---measurement model of Fig.~\ref{fig:8view_results},\ref{fig:4view_results},\ref{fig:2view_results}, (c) compressed sensing MRI (CS-MRI)---measurement model of Fig.~\ref{fig:mri_results}.}
    \vspace{-0.5cm}
    \label{fig:forward_model}
\end{figure*}
%

%%%%%%%%% BODY TEXT
\section{Introduction}
\label{sec:intro}

Diffusion models learn the data distribution implicitly by learning the gradient of the log density (i.e. $\nabla_\xb \log p_{\rm{data}}(\xb)$; score function)~\cite{ho2020denoising,song2020score}, which is used at inference to create generative samples. These models are known to generate high-quality samples, cover the modes well, and be highly robust to train, as it amounts to merely minimizing a mean squared error loss on a denoising problem. Particularly, diffusion models are known to be much more robust than other popular generative models~\cite{dhariwal2021diffusion}, for example, generative adversarial networks (GANs). Furthermore, one can use pre-trained diffusion models to solve inverse problems in an unsupervised fashion~\cite{song2020score,chung2022come,chung2022score,chung2022improving,kawar2022denoising}. Such strategies has shown to be highly effective in many cases, often establishing the new state-of-the-art on each task. Specifically, applications to sparse view computed tomography (SV-CT)~\cite{song2022solving,chung2022improving}, compressed sensing MRI (CS-MRI)~\cite{chung2022score,chung2022come,song2022solving}, super-resolution~\cite{choi2021ilvr,chung2022come,kawar2022denoising}, inpainting~\cite{kawar2022denoising,chung2022improving} among many others, have been proposed.

Nevertheless, to the best of our knowledge, all the methods considered so far focused on 2D imaging situations. This is mostly due to the high-dimensional nature of the generative constraint. Specifically, diffusion models generate samples by starting from pure noise, and iteratively denoising the data until reaching the clean image. Consequently, the generative process involves staying in the {\em same} dimension as the data, which is prohibitive when one tries to scale the data dimension to 3D. 
One should also note that training a 3D diffusion model amounts to learning the 3D prior of the data density. This is undesirable in two aspects. First, the model is data hungry, and hence training a 3D model would typically require thousands of {\em volumes}, compared to 2D models that could be trained with less than 10 volumes. Second, the prior would be needlessly complicated: when it comes to dynamic imaging or 3D imaging, exploiting the spatial/temporal correlation~\cite{sun2019exploiting,jung2009k} is standard practice. Naively modeling the problem as 3D would miss the chance of leveraging such information.

Another much more well-established method for solving 3D inverse problems is model-based iterative reconstruction (MBIR)~\cite{katsura2012model,liu2014model}, where the problem is formulated as an optimization problem of weighted least squares (WLS), constructed with the data consistency term, and the regularization term. One of the most widely acknowledged regularization in the field is the total variation (TV) penalty~\cite{sidky2008image,liu2012adaptive}, known for its intriguing properties: edge-preserving, while imposing smoothness. While the TV prior has been widely explored, it is known to fall behind the data-driven prior of the modern machine learning practice, as the function is too simplistic to fully model how the image ``looks like''.

In this work, we propose DiffusionMBIR, a method to combine the best of both worlds: we incorporate the MBIR optimization strategy into the diffusion sampling steps in order to {\em augment} the data-driven prior with the conventional TV prior, imposed to the $z$-direction only. Particularly, the standard reverse diffusion (i.e. denoising) steps are run independently with respect to the $z$-axis, and hence standard 2D diffusion models can be used. {Subsequently, the data consistency step is imposed by aggregating the slices, then taking a single update step of the alternating direction method of multipliers (ADMM)~\cite{boyd2011distributed}.} This step effectively coerces the cross-talk between the slices with the measurement information, and the TV prior. For efficient optimization, we further propose a strategy in which we call {\em variable sharing}, which enables us to only use a {\em single} sweep of ADMM and conjugate gradient (CG) per denoising iteration. Note that our method is fully general in that we are not restricted to the given forward operator at test time. Hence, we verify the efficacy of the method by performing extensive experiments on sparse-view CT (SV-CT), limited angle CT (LA-CT), and compressed sensing MRI (CS-MRI): out method shows consistent improvements over the current diffusion model-based inverse problem solvers, and shows strong performance on {\em all} tasks (For representative results, see Fig.~\ref{fig:cover}. For conceptual illustration of the inverse problems, see Fig.~\ref{fig:forward_model}).

In short, the main contributions of this paper is to devise a diffusion model-based reconstruction method that 1) operate with the voxel representation, 2) is memory-efficient such that we can scale our solver to much higher dimensions (i.e. $> 256^3$), and 3) is not data hungry, such that it can be trained with less than ten 3D volumes.

%-------------------------------------------------------------------------
\section{Background}
\label{sec:background}

\noindent
\textbf{Model-based iterative reconstruction (MBIR).~}
\label{sec:mbir}
Consider a linear forward model for an imaging system (e.g. CT, MRI)
\begin{align}
\label{eq:forward}
    \yb = \Ab\xb + \nb,
\end{align}
where $\yb \in \Rd^m$ is the measurement (i.e. sinogram, k-space), $\xb \in \Rd^n$ is the image that we wish to reconstruct, $\Ab \in \Rd^{m \times n}$ is the discrete transform matrix (i.e. Radon, Fourier\footnote{While we denote real-valued transforms and measurements for the simplicity of exposition, the discrete Fourier transform (DFT) matrix, and the corresponding measurement are complex-valued.}), and $\nb$ is the measurement noise in the system. As the problem is ill-posed, a standard approach for the inverse problem
that estimates the unknown image $\xb$ from the measurement $\yb$ is to perform the following regularized reconstruction:
\begin{align}
\label{eq:wls}
    \xb^* = \argmin_{\xb} \frac{1}{2}\|\yb - \Ab\xb\|_2^2 + R(\xb),
\end{align}
where $R$ is the suitable regularization for $\xb$, for instance, sparsity in some transformed domain. One widely used function is the TV penalty, $R(\xb) = \|\Db\xb\|_{2,1}$, where $\Db := [\Db_x, \Db_y, \Db_z]^T$ computes the finite difference in each axis. Minimization of \eqref{eq:wls} can be performed with robust optimization algorithms, such as fast iterative soft thresholding algorithm (FISTA)~\cite{beck2009fast} or ADMM.

\noindent
\textbf{Score-based diffusion models.~}
\label{sec:diffusion}
A score-based diffusion model is a generative model that defines the generative process as the {\em reverse} of the data {\em noising} process. Specifically,
consider the stochastic process $\{\xb(t) \triangleq \xb_t\},\, t \in [0, 1]$, where we introduce the {\em time} variable $t$ to represent the evolution of the random variable. Particularly, we define $p(\xb_0) \triangleq p_{\rm{data}}(\xb)$, i.e. the data distribution, and $p(\xb_T)$ to approximately a Gaussian distribution. The evolution can be formalized with the following stochastic differential equation
\begin{align}
\label{eq:forward_sde}
    d\xb = \fb(\xb, t)\,dt + g(t)\,d\wb,
\end{align}
where $\fb(\xb, t): \Rd^{n \times 1} \mapsto \Rd^n$ is the drift function, $g(t): \Rd \mapsto \Rd$ is the scalar diffusion function, and $\wb$ is the $n-$dimensional standard Brownian motion~\cite{sarkka2019applied}. Let $\fb(\xb, t) = 0, g(t) = \sqrt{\frac{d[\sigma^2(t)]}{dt}}$. Then, the SDE simplifies to the following Brownian motion
\begin{align}
\label{eq:ve_sde}
    d\xb = \sqrt{\frac{d[\sigma^2(t)]}{dt}}\,d\wb,
\end{align}
in which the mean remains the same across the evolution, while Gaussian noise will be continuously added to $\xb$, eventually approaching pure Gaussian noise as the noise term dominates. This is the so called variance-exploding SDE (VE-SDE) in the literature~\cite{song2020score}, and as we construct all our methods on the VE-SDE, we derive what follows from \eqref{eq:ve_sde}. Directly applying Anderson's theorem~\cite{anderson1982reverse,song2020score} leads to the following reverse SDE
\begin{align}
\label{eq:ve_sde_reverse}
    d\xb = -\frac{d[\sigma^2(t)]}{dt} \nabla_{\xb_t} \log p(\xb_t)\,dt + \sqrt{\frac{d[\sigma^2(t)]}{dt}}\,d\bar{\wb},
\end{align}
where $dt, d\bar{\wb}$ are the reverse time differential, and the reverse standard $n-$dimensional Brownian motion. \eqref{eq:ve_sde_reverse} defines the {\em generative} process of the diffusion model, where the equation can be solved by numerical integration. Notably, the key workhorse in the integration step is the score function $\nabla_{\xb_t} \log p(\xb_t)$, that can be trained with denoising score matching (DSM)~\cite{vincent2011connection}
\begin{align}
\label{eq:dsm}
    \min_\theta \Ed_{t,\xb(t)}\left[\lambda(t)\|\s_\theta(\xb(t), t) - \nabla_{\xb_t}\log p(\xb(t)|\xb(0))\|_2^2\right],
\end{align}
where $\s_\theta(\xb(t), t): \Rd^{n \times 1} \mapsto \Rd^n$ is a time-dependent neural network, and $\lambda(t)$ is the weighting scheme. Since $\nabla_{\xb_t}\log p(\xb(t)|\xb(0))$ is simply the {\em residual noise} added to $\xb(t)$ from $\xb(0)$ scaled with noise variance, optimizing for \eqref{eq:dsm} amounts to training a residual denoiser across multiple noise scales - a fairly robust training scheme. While the training is robust, the equivalence between DSM and explicit score matching (ESM) can be established~\cite{vincent2011connection} in the optimization sense, and hence $s_{\theta^*}(\xb(t),t) \simeq \nabla_{\xb_t} \log p(\xb_t)$ can be used as a plug-in approximate in practice, i.e.
\begin{align}
\label{eq:ve_sde_reverse_stheta}
    d\xb \simeq -\frac{d[\sigma^2(t)]}{dt} \s_{\theta^*}(\xb(t), t)\,dt + \sqrt{\frac{d[\sigma^2(t)]}{dt}}\,d\bar{\wb}.
\end{align}
One can solve \eqref{eq:ve_sde_reverse_stheta} with, e.g. the predictor-corrector (PC) sampler~\cite{song2020score} by discretization of the time interval $[0, 1]$ to $N$ bins.

\noindent
\textbf{3D diffusion.~}
\label{sec:3d_diffusion}
The generative process (i.e. reverse diffusion), explicitly represented by \eqref{eq:ve_sde_reverse_stheta}, runs in the full data dimension $\Rd^n$. It is widely known that the voxel representation of 3D data is heavy, and scaling the data size over $64^3$ requires excessive GPU memory~\cite{liu2019point}. For example, a recent work that utilizes diffusion models for 3D shape reconstruction~\cite{waibel2022diffusion} uses the dataset of size $64^3$. Other works that use diffusion models for 3D generative modeling typically focus on the more efficient point cloud representation~\cite{luo2021diffusion,zhou20213d,lyu2021conditional}, where the number of point clouds remain less than a few thousand (e.g. $2048 \ll 64^3$ in~\cite{luo2021diffusion,zhou20213d}). Naturally, point cloud representations are efficient but extremely sparse, certainly not suitable for the problem of tomographic reconstruction, where we require accurate estimation of the interior. 

There is one concurrent workshop paper that aims for designing a diffusion model that can model the 3D voxel representation~\cite{pinaya2022brain}. In~\cite{pinaya2022brain}, the authors train a score function that can model $160\times224\times160$ volumes, by training a latent diffusion model~\cite{rombach2022high}, where the latent dimension is relatively small (i.e. $20\times28\times20$). However, the model requires 1000 synthetic 3D volumes for the training dataset. More importantly, using {\em latent} diffusion models for solving inverse problems is not straightforward, and has never been reported in literature.

\noindent
\textbf{Solving inverse problems with diffusion.~}
\label{sec:solving_ip_diff}
Solving the reverse SDE with the approximated score function \eqref{eq:ve_sde_reverse_stheta} amounts to sampling from the prior distribution $p(\xb)$. For the case of solving inverse problem, we desire to sample from the posterior distribution $p(\xb|\yb)$, where the relationship between the two can be formulated by the Bayes' rule $p(\xb|\yb) = p(\xb)p(\yb|\xb)/p(\yb)$, leading to
\begin{align}
\label{eq:grad_log_bayes}
    \nabla_{\xb}\log p(\xb|\yb) = \nabla_{\xb}\log p(\xb) + \nabla_{\xb}\log p(\yb|\xb).
\end{align}
Here, the likelihood term enforces the data consistency, and thereby inducing samples that satisfy $\yb = \Ab\xb$. Two approaches to incorporating \eqref{eq:grad_log_bayes} exist in the literature. First, one can split the update step into the 1) prior update (i.e. denoising), and then 2) projection in to the measurement subspace~\cite{song2020score,chung2022come}. Formally, in the discrete setting,
\begin{align}
\label{eq:alt_denoise}
    \xb'_{i-1} &\gets {\rm Solve}(\xb_{i-1}, \s_{\theta^*}),\\
    \xb_{i} &\gets \Pc_{\{\xb|\Ab\xb = \yb\}}(\xb'_{i-1}),
\label{eq:alt_proj}
\end{align}
where ${\rm Solve}$ denotes a general numerical solver that can solve the reverse-SDE in \eqref{eq:ve_sde_reverse_stheta}, and $\Pc_C$ denotes the projection operator to the set $C$. Specifically, when using the Euler-Maruyama discretization, the equation reads\footnote{For all equations and algorithms that are presented, we refer to the sampled random Gaussian noise as $\bm{\epsilon} \sim \Nc(0,\Ib)$, unless specified otherwise.}
\begin{align}
\label{eq:alt_denoise_ex}
    \xb'_{i-1} &\gets (\sigma_{i}^2 - \sigma_{i-1}^2)\s_{\theta^*}(\xb_{i-1},i-1) + \sqrt{\sigma_i^2 - \sigma_{i-1}^2}\bm{\epsilon},\\
    \xb_{i} &\gets \Pc_{\{\xb|\Ab\xb = \yb\}}(\xb'_{i-1}).
\label{eq:alt_proj_ex}
\end{align}
Note that the stochasticity of ${\rm Solve(\cdot)}$ is implicitly defined.
It was shown in~\cite{song2020score} that using the PC solver, which alternates between the numerical SDE solver and monte carlo markov chain (MCMC) steps leads to superior performance. Throughout the manuscript, we refer to a single step of PC sampler as ${\rm Solve(\cdot)}$ unless specified otherwise.
Alternatively, one can try to explicitly approximate the gradient of the log likelihood and take the update in a single step~\cite{chung2022improving}.
%}

%------------------------------------------------------------------------
\section{DiffusionMBIR}
\label{sec:main_contributions}

\subsection{Main idea}

\begin{figure*}[!t]
    \centering
    \includegraphics[width=1.0\textwidth]{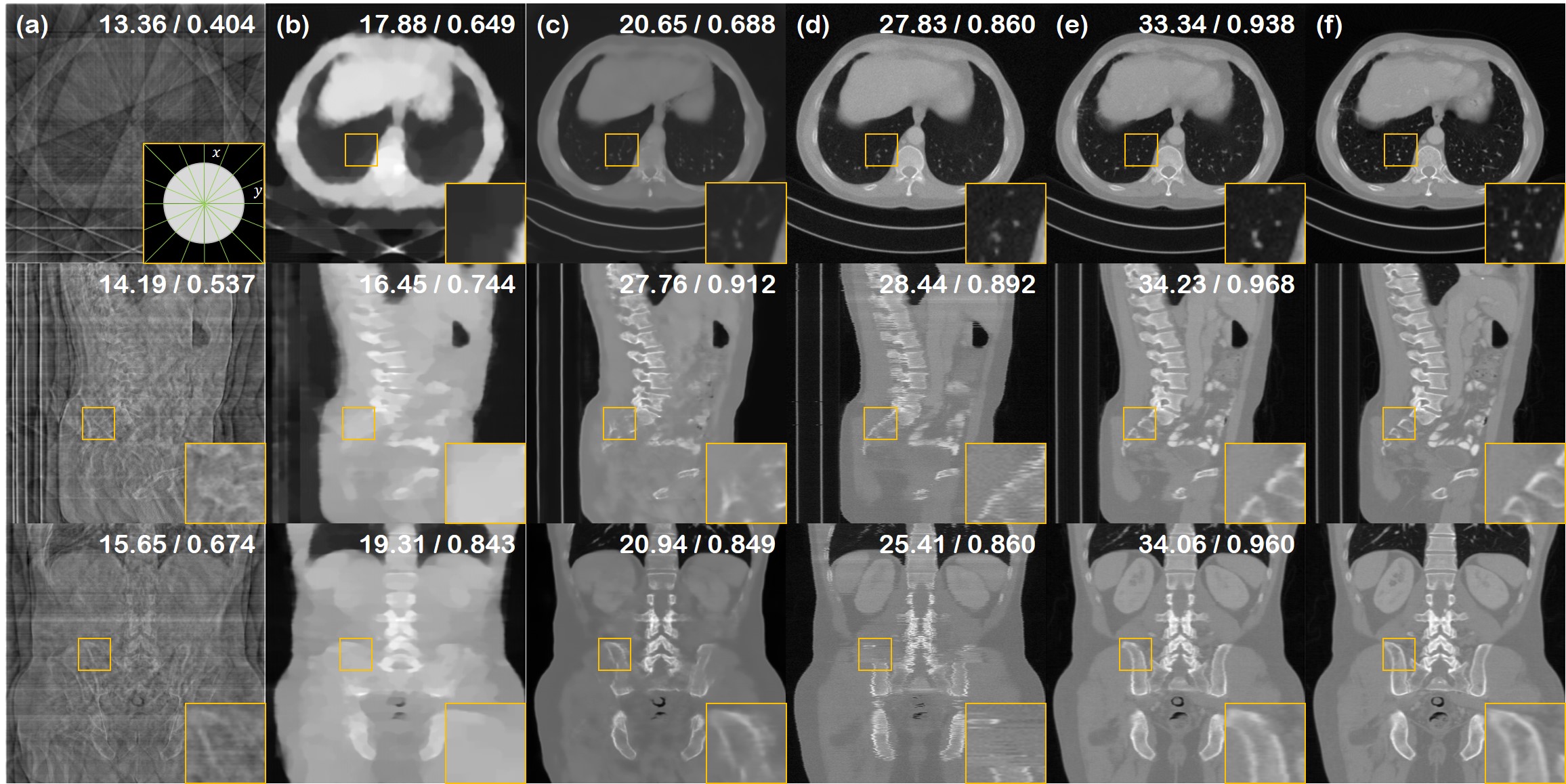}
    \caption{8-view SV-CT reconstruction results of the test data (First row: axial slice, second row: sagittal slice, third row: coronal slice). (a) FBP, (b) ADMM-TV, (c) Lahiri {\em et al.}~\cite{lahiri2022sparse}, (d) Chung {\em et al.}~\cite{chung2022improving}, (e) proposed method, (f) ground truth. PSNR/SSIM values presented in the upper right corner. Green lines in the inset of first row (a): measured angles.}
    \label{fig:8view_results}
\end{figure*}

To efficiently utilize the diffusion models for 3-D reconstruction, one possible solution would be to apply 2-D diffusion models slice by slice.
Specifically, \eqref{eq:alt_denoise},\eqref{eq:alt_proj} could be applied {\em parallel} with respect to the $z-$axis. 
However, this approach has one fundamental limitation.
When the steps are run without considering the inter-dependency between the slices, the slices that are reconstructed will not be coherent with each other (especially when we have sparser view angles). Consequently, when viewed from the coronal/sagittal slice, the images contain severe artifacts.(see Fig.~\ref{fig:8view_results},\ref{fig:lact_results} (d) row 2-3).

In order to address this issue, we are interested in combining the advantages from the MBIR and the diffusion model to oppress unwanted artifacts.
Specifically, our proposal is to adopt the alternating minimization approach in \eqref{eq:alt_denoise},\eqref{eq:alt_proj}, but rather than applying them in 2-D domain,
the diffusion-based denoising step in \eqref{eq:alt_denoise} is applied slice-by-slice,
whereas the 2-D projection step in \eqref{eq:alt_proj} is replaced with
the ADMM update step in 3-D volume. Specifically, we consider the following sub-problem
\begin{align}
\label{eq:wls_tvz}
    \min_\xb \frac{1}{2}\|\yb - \Ab\xb\|_2^2 + \|\Db_z \xb\|_1, 
\end{align}
where unlike the conventional TV algorithms that take $\|\Db \xb\|_1$, we only take the $\ell_1$ norm of the finite difference in the $z-$axis. This choice stems from the fact that the prior with respect to the $xy$ plane is already taken care of with the neural network $\s_{\theta^*}$, and all we need to imply is the spatial correlation with respect to the remaining direction. In other words, we are augmenting the generative prior with the model-based sparsity prior. From our experiments, we observe that our prior augmentation strategy is highly effective in producing coherent 3D reconstructions throughout all the three axes.

\subsection{Algorithmic steps}
We arrive at the update steps
\begin{align}
    \xb^{+} &= (\Ab^T\Ab + \rho \Db_z^T\Db_z)^{-1}(\Ab^T\yb + \rho \Db^T(\zb - \wb))\label{eq:admm_x}\\
    \zb^{+} &= \Sc_{\lambda/\rho}(\Db_z \xb^{+} + \wb)\label{eq:admm_z}\\
    \wb^{+} &= \wb + \Db_z \xb^{+} - \zb^{+},
    \label{eq:admm_w}
\end{align}
where $\rho$ is the hyper-paremeter for the method of multipliers, and $\Sc$ is the soft thresholding operator. Moreover, \eqref{eq:admm_x} can be solved with conjugate gradient (CG), which efficiently finds a solution for $\xb$ that satisfies $\Ab\xb = \bb$: we denote running $K$ iterations of CG with initial point $\xb$ as ${\code{CG}}(\Ab, \bb, {\xb}, K)$.
Full derivation for the ADMM steps is provided in Supplementary section~\ref{sec:admm_tv}. For simplicity, we denote one sweep of \eqref{eq:admm_x},\eqref{eq:admm_z},\eqref{eq:admm_w} as $\xb^+,\zb^+,\wb^+ = {\code{ADMM}} (\xb,\zb,\wb)$. Iterative application of ${\code{ADMM}} (\xb,\zb,\wb)$ would robustly solve the minimization problem in \eqref{eq:wls_tvz}. Hence, the naive implementation of the proposed algorithm would be
\begin{align}
\label{eq:DiffusionMBIR-denoise}
    \xb'_{i-1} &\gets {\rm Solve}(\xb_i, \s_{\theta^*}),\\
    \xb_{i-1} &\gets \argmin_{\xb'_{i-1}} \frac{1}{2}\|\yb - \Ab\xb'_{i-1}\|_2^2 + \|\Db_z \xb'_{i-1}\|_1. 
\label{eq:DiffusionMBIR-admm}
\end{align}
Specifically, \eqref{eq:DiffusionMBIR-denoise} would amount to parallel denoising for each slice, whereas \eqref{eq:DiffusionMBIR-admm} augments the $z$-directional
TV prior and impose consistency. See the detailed  (slow version) solving steps in Algorithm~\ref{alg:diffusion_mbir_slow} of the supplementary section. 
Here, note that there are three sources of iteration in the algorithm: 1) Numerical integration of SDE, indexed with $i$, 2) $\rm{ADMM}$ iteration, and 3) the inner CG iteration, used to solve \eqref{eq:admm_x}. 

Since diffusion models are slow in itself, multiplicative additional cost of factors 2),3) will be prohibitive, and should be refrained from. In the following, we devise a simple method to reduce this cost dramatically.

\begin{algorithm}[!t]
\caption{DiffusionMBIR (fast; variable sharing)}
\begin{algorithmic}[1]
\Require $s_{\theta^*}, N, \lambda, \rho, \{\sigma_i\}$
\State $\xb_N \sim \Nc({\bf 0}, \sigma_{T}^{2} \Ib)$
\State $\zb_N \gets$ \code{torch.zeros\_like}$({\xb_N})$
\State $\wb_N \gets$ \code{torch.zeros\_like}$({\xb_N})$
\For{$i = N-1:0$} \Comment{SDE iteration} \do \\
\State {$\xb'_{i} \gets {\rm Solve}(\xb_{i+1}, s_{\theta^*})$}
\State $\Ab_{\rm{CG}} \gets \Ab^T\Ab + \rho \Db_z^T\Db_z$
\State $\bb_{\code{CG}} \gets \Ab^T\yb + \rho \Db_z^T(\zb_{i+1} - \wb_{i+1})$
\State $\xb_i \gets {\code{CG}}(\Ab_{\code{CG}}, \bb_{\code{CG}}, {\xb'_i}, 1)$
\State $\zb_{i} \gets \Sc_{\lambda/\rho}(\Db_z\xb_i + \wb_{i+1})$
\State $\wb_{i} \gets \wb_{i+1} + \Db_z\xb_i - \zb_i$
\EndFor
\State \textbf{return} $\xb_0$
\end{algorithmic}\label{alg:diffusion_mbir_fast}
\end{algorithm}

\begin{table*}[!t]
\centering
\setlength{\tabcolsep}{0.2em}
\resizebox{1.0\textwidth}{!}{% <------ Don't forget this %
\begin{tabular}{lllllll@{\hskip 15pt}llllll@{\hskip 15pt}llllll}
\toprule
{} & \multicolumn{6}{c}{\textbf{8-view}} & \multicolumn{6}{c}{\textbf{4-view}} & \multicolumn{6}{c}{\textbf{2-view}} \\
\cmidrule(lr){2-7}
\cmidrule(lr){8-13}
\cmidrule(lr){14-19}
{} & \multicolumn{2}{c}{\textbf{Axial$^*$}} & \multicolumn{2}{c}{\textbf{Coronal}} & \multicolumn{2}{c}{\textbf{Sagittal}} & \multicolumn{2}{c}{\textbf{Axial$^*$}} & \multicolumn{2}{c}{\textbf{Coronal}} & \multicolumn{2}{c}{\textbf{Sagittal}} & \multicolumn{2}{c}{\textbf{Axial$^*$}} & \multicolumn{2}{c}{\textbf{Coronal}} & \multicolumn{2}{c}{\textbf{Sagittal}} \\
\cmidrule(lr){2-3}
\cmidrule(lr){4-5}
\cmidrule(lr){6-7}
\cmidrule(lr){8-9}
\cmidrule(lr){10-11}
\cmidrule(lr){12-13}
\cmidrule(lr){14-15}
\cmidrule(lr){16-17}
\cmidrule(lr){18-19}
{\textbf{Method}} & {PSNR $\uparrow$} & {SSIM $\uparrow$} & {PSNR $\uparrow$} & {SSIM $\uparrow$} & {PSNR $\uparrow$} & {SSIM $\uparrow$} & {PSNR $\uparrow$} & {SSIM $\uparrow$} & {PSNR $\uparrow$} & {SSIM $\uparrow$} & {PSNR $\uparrow$} & {SSIM $\uparrow$} & {PSNR $\uparrow$} & {SSIM $\uparrow$} & {PSNR $\uparrow$} & {SSIM $\uparrow$} & {PSNR $\uparrow$} & {SSIM $\uparrow$} \\
\midrule
DiffusionMBIR~\textcolor{trolleygrey}{(ours)} & \textbf{33.49} & \textbf{0.942} & \textbf{35.18} & \textbf{0.967} & \textbf{32.18} & \textbf{0.910} & \textbf{30.52} & \textbf{0.914} & \textbf{30.09} & \textbf{0.938} & \textbf{27.89} & \textbf{0.871} & \underline{24.11} & \underline{0.810} & \underline{23.15} & \textbf{0.841} & \textbf{21.72} & \textbf{0.766}\\
\cmidrule(l){1-19}
Chung {\em et al.}~\cite{chung2022improving} & \underline{28.61} & \underline{0.873} & \underline{28.05} & \underline{0.884} & \underline{24.45} & \underline{0.765} & \underline{27.33} & \underline{0.855} & \underline{26.52} & \underline{0.863} & \underline{23.04} & \underline{0.745} & \textbf{24.69} & \textbf{0.821} & \textbf{23.52} & \underline{0.806} & \underline{20.71} & \underline{0.685}\\
Lahiri {\em et al.}~\cite{lahiri2022sparse} & 21.38 & 0.711 & 23.89 & 0.769 & 20.81 & 0.716 & 20.37 & 0.652 & 21.41 & 0.721 & 18.40 & 0.665 & 19.74 & 0.631 & 19.92 & 0.720 & 17.34 & 0.650\\
FBPConvNet~\cite{jin2017deep} & 16.57 & 0.553 & 19.12 & 0.774 & 18.11 & 0.714 & 16.45 & 0.529 & 19.47 & 0.713 & 15.48 & 0.610 & 16.31 & 0.521 & 17.05 & 0.521 & 11.07 & 0.483\\
ADMM-TV~\cite{jin2017deep} & 16.79 & 0.645 & 18.95 & 0.772 & 17.27 & 0.716 & 13.59 & 0.618 & 15.23 & 0.682 & 14.60 & 0.638 & 10.28 & 0.409 & 13.77 & 0.616 & 11.49 & 0.553\\
\bottomrule
\end{tabular}
}
\vspace{-0.2cm}
\caption{
Quantitative evaluation of SV-CT (8, 4, 2-view) (PSNR, SSIM) on the AAPM 256$\times$256 test set. \textbf{Bold}: Best, \underline{under}: second best.$^*$: the plane where the diffusion model prior takes place. Holds the same for Table.~\ref{tab:la-ct},\ref{tab:cs-mri}.$^*$: the plane where the diffusion model prior takes place. Holds the same for Table.~\ref{tab:la-ct},\ref{tab:cs-mri}
}
\vspace{-0.5cm}
\label{tab:8view_quantitative}
\end{table*}

%\noindent
\paragraph{Fast and efficient implementation (variable sharing)}

In Algorithm~\ref{alg:diffusion_mbir_slow}, we re-initialize the primal variable $\zb$, and the dual variable $\wb$, everytime before the ADMM iteration runs for the $i^{\text{th}}$ iteration of the SDE. In turn, this would lead to slow convergence of the ADMM algorithm, as burn-in period for the variables $\zb,\wb$ would be required for the first few iterations. Moreover, since solving for diffusion models would have large number of discretization steps $N$, the difference between the two adjacent iterations $\xb_i$ and $\xb_{i+1}$ is minimal. When dropping the values of $\zb, \wb$ from the $i+1^{\text{th}}$ iteration and re-initializing at the $i^{\text{th}}$ iteration, one would be dropping valuable information, and wasting compute. Hence, we propose to initialize both $\zb_N, \wb_N$ as a {\em global} variable before the start of the SDE iteration, and keep the updated values throughout. Interesting enough, we find that choosing $M=1, K=1$, i.e. {\em single} iteration for both ADMM and CG are necessary for high fidelity reconstruction. Our fast version of DiffusionMBIR is presented in Algorithm~\ref{alg:diffusion_mbir_fast}.

Another caveat is when running the neural network forward pass through the entire volume is not feasible memory-wise, for example, when fitting the solver into a single commodity GPU. One can circumvent this by dividing the batch dimension\footnote{In our implementation, the batch dimension corresponds to the $z-$axis, as 2D slices are stacked.} into sub-batches, running the denoising step for the sub-patches separately, and then aggregating them into the full volume again. The \code{ADMM} step can be applied to the full volume after the aggregation, which would yield the same solution with Algorithm~\ref{alg:diffusion_mbir_fast}. For both the slow and the fast version of the algorithm, one can also apply a projection to the measurement subspace at the end when one wishes to exactly match the measurement constraint.
\vspace{-0.3cm}

%-------------------------------------------------------------------------
\section{Experimental setup}

\begin{figure*}[!t]
    \centering
    \includegraphics[width=1.0\textwidth]{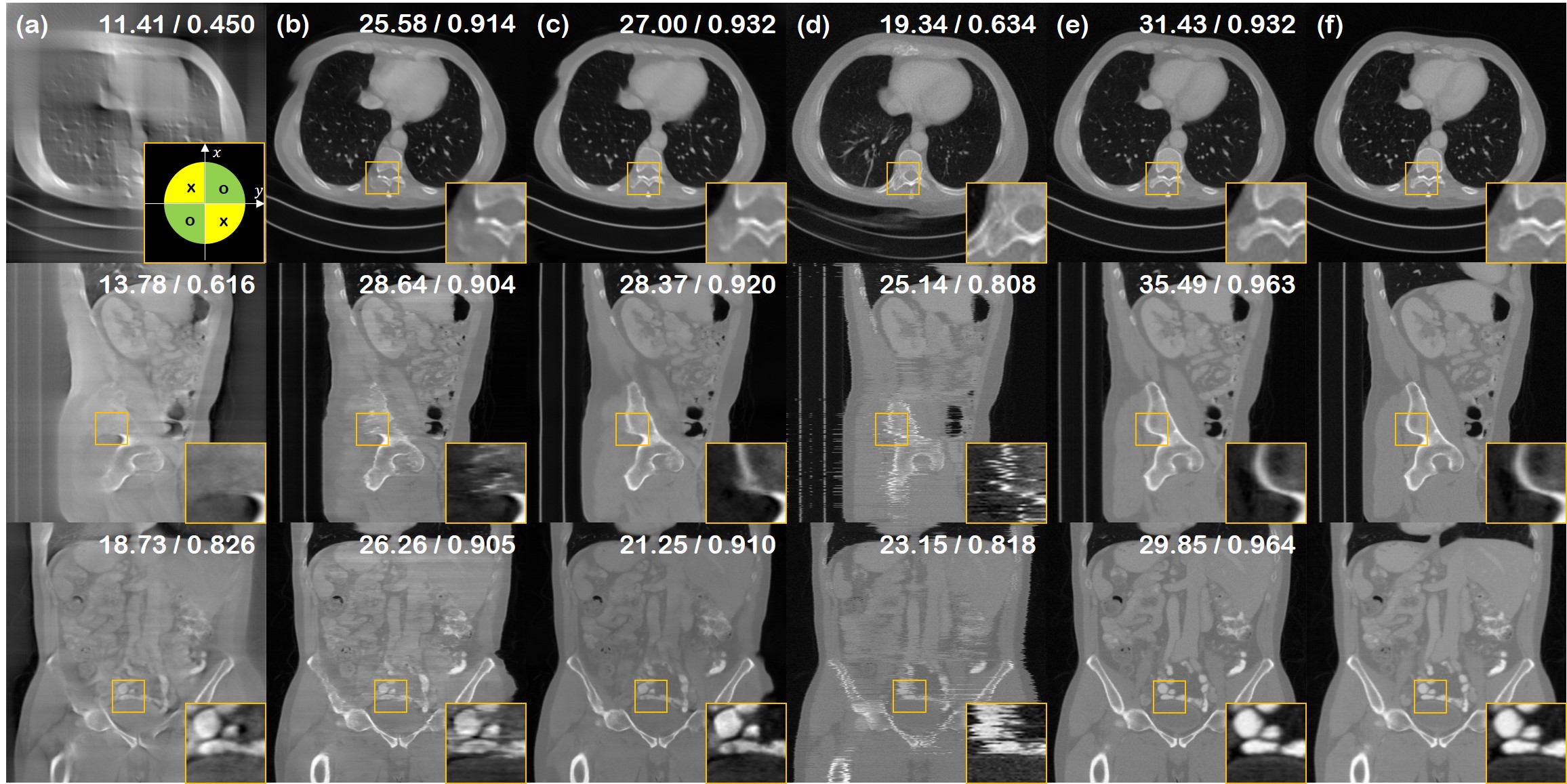}
    \caption{90$^\circ$ LA-CT reconstruction results of the test data (First row: axial slice, second row: sagittal slice, third row: coronal slice). (a) FBP, (b) Zhang {\em et al.}~\cite{zhang2016image}, (c) Lahiri {\em et al.}~\cite{lahiri2022sparse}, (d) Chung {\em et al.}~\cite{chung2022improving}, (e) proposed method, (f) ground truth. PSNR/SSIM values presented in the upper right corner. Green area in the inset of first row (a): measured, Yellow area in the inset of first row (a): not measured.}
    \vspace{-0.5cm}
    \label{fig:lact_results}
\end{figure*}

We conduct experiments on three most widely studied tasks in medical image reconstruction: 1) sparse view CT (SV-CT), 2) limited angle CT (LA-CT), and 3) compressed sensing MRI (CS-MRI). Specific details can be found in the supplementary section~\ref{sec:supp_exp_details}.

\noindent
\textbf{Dataset.~}
For both CT reconstruction tasks (i.e. SV-CT, LA-CT) we use the data from the AAPM 2016 CT low-dose grand challenge. All volumes except for one are used for training the 2D score function, and one volume is held-out for testing. For the task of CS-MRI, we take the data from the multimodal brain tumor image segmentation benchmark (BRATS)~\cite{menze2014multimodal} 2018 FLAIR volume for testing. Note that we use a pre-trained score function that was trained on fastMRI knee~\cite{zbontar2018fastmri} images only, and hence we need not split the train/test data here.

\noindent
\textbf{Network training, inference.~}
For CT tasks, we train the \code{ncsnpp} model~\cite{song2020score} on the AAPM dataset which consists of about 3000 2D slices of training data.
For the CS-MRI task, we take the pre-trained model checkpoint from\footnote{\url{https://github.com/HJ-harry/score-MRI}}~\cite{chung2022score}. For inference (i.e. generation; inverse problem solving), we base our sampler on the predictor-corrector (PC) sampling scheme of~\cite{song2020score}. We set $N = 2000$, which amounts to 4000 iterations of neural function evaluation with $\s_{\theta^*}$.

\noindent
\textbf{Comparison methods and evaluation.~}
For CT tasks, we first compare our method with Chung {\em et al.}~\cite{chung2022improving}, which is another diffusion model approach for CT reconstruction, outperforming~\cite{song2021solving}. As using the manifold constrained gradient (MCG) of~\cite{chung2022improving} requires ~10GB of VRAM for a {\em single} 2D slice (256$\times$256), it is infeasible for us to leverage such gradient step for our 3D reconstruction. Thus, we employ the projection onto convex sets (POCS) strategy of~\cite{chung2022improving}, which amounts to taking algebraic reconstruction technique (ART) in each data-consistency imposing step.
We also compare against some of the best-in-class fully supervised methods. Namely, we include Lahiri {\em et al.}~\cite{lahiri2022sparse}, and FBPConvNet~\cite{jin2017deep} (SV-CT) / Zhang {\em et al.}~\cite{zhang2016image} (LA-CT) as baselines. For the implementation of~\cite{lahiri2022sparse}, we use 2 stages, but implement the De-streaking CNN as U-Nets rather than simpler CNNs. Finally, we include isotropic ADMM-TV, which uses the regularization function $R(\xb) = \|\Db\xb\|_{2,1}$.

In the CS-MRI experiments, we compare against score-MRI~\cite{chung2022score} as a representative diffusion model-based solver. Moreover, we include comparisons with DuDoRNet~\cite{zhou2020dudornet}, and U-Net~\cite{zbontar2018fastmri}. Note that all networks including ours, the training dataset (fastMRI knee) was set deliberately different from the testing data (BRATS Flair), as it was shown that diffusion model-based inverse problem solvers are fairly robust to out of distribution (OOD) data in CS-MRI settings~\cite{chung2022score,jalal2021robust}.
Quantitative evaluation was performed with two standard metrics: peak-signal-to-noise-ratio (PSNR), and structural similarity index (SSIM). We report on metrics that are averaged over each planar direction, as we expect different performance for $xy$-slices as compared to $xz$- and $yz$-slices.
\vspace{-0.2cm}

\section{Results}
\label{sec:results}
In this section, we present the results of the proposed method. For further experiments and ablation studies, see supplementary section~\ref{sec:supp_experiments}.

\noindent
\textbf{Sparse-view CT.}
We present the quantitative metrics of the SV-CT reconstruction results in Table~\ref{tab:8view_quantitative}. The table shows that the proposed method outscores the baselines by large margins in most of the settings. Fig.~\ref{fig:8view_results} and Supplementary Fig.~\ref{fig:4view_results} show the 8,4-view SV-CT reconstruction result. As shown at the first row of each figure, axial slices of the proposed method have restored much finer details compared to the baselines. Furthermore, the results of sagittal and coronal slices in the second and third rows imply that DiffusionMBIR could maintain the structural connectivity of the original structures in all directions. 
In contrast, Chung et al.~\cite{chung2022improving} performs well on reconstructing the axial slices, but do not have spatial integrity across the $z$ direction, leading to shaggy artifacts that can be clearly seen in coronal/sagittal slices. Lahiri et al.~\cite{lahiri2022sparse} often omits important details, and is not capable of reconstruction, especially when we only have 4 number of views. ADMM-TV hardly produces satisfactory results due to the extremely limited setting.

\begin{figure*}[]
    \centering
    \includegraphics[width=1.0\textwidth]{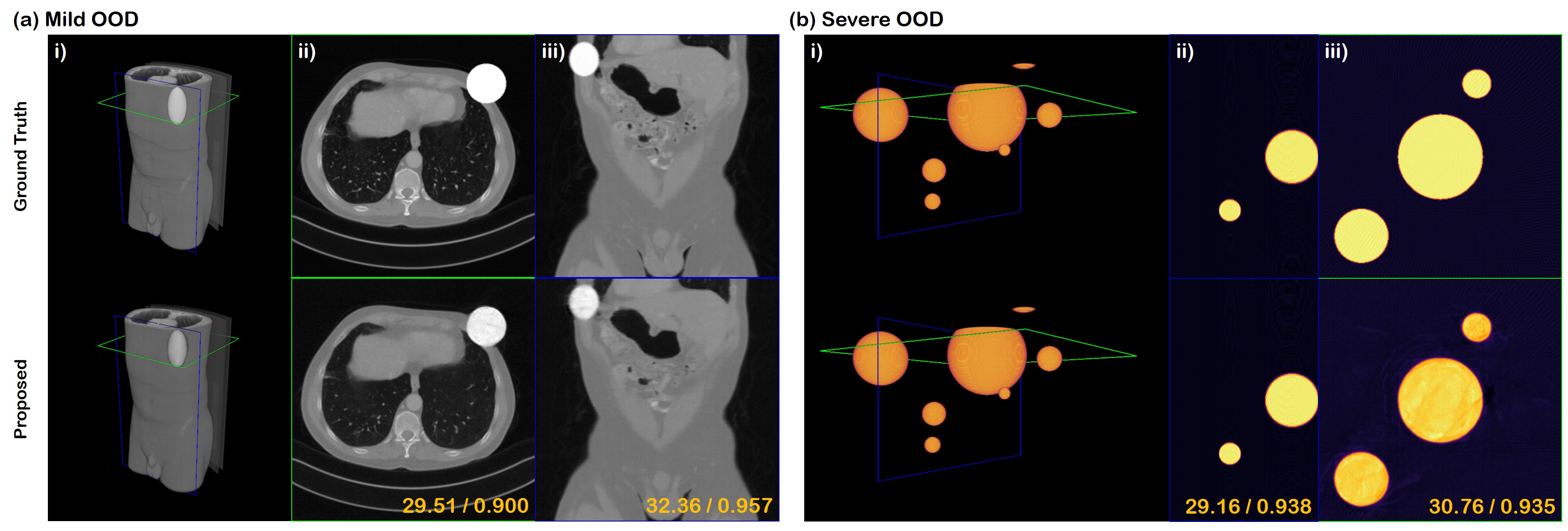}
    \vspace{-0.5cm}
    \caption{8-view SV-CT reconstruction results of the OOD data (Same geometry as in Fig.~\ref{fig:8view_results}). (a) Ellipsis laid on top of the test data volume, (b) Phantom that consists of spheres located randomly.}
    \label{fig:ood}
    \vspace{-0.5cm}
\end{figure*}

\noindent
\textbf{Limited angle CT.~}
\begin{table}[t]
\centering
\setlength{\tabcolsep}{0.2em}
\resizebox{0.45\textwidth}{!}{% <------ Don't forget this %
\begin{tabular}{lllllll}
\toprule
{} & \multicolumn{2}{c}{\textbf{Axial$^*$}} & \multicolumn{2}{c}{\textbf{Coronal}} & \multicolumn{2}{c}{\textbf{Sagittal}} \\
\cmidrule(lr){2-3}
\cmidrule(lr){4-5}
\cmidrule(lr){6-7}
{\textbf{Method}} & {PSNR $\uparrow$} & {SSIM $\uparrow$} & {PSNR $\uparrow$} & {SSIM $\uparrow$} & {PSNR $\uparrow$} & {SSIM $\uparrow$}\\
\midrule
DiffusionMBIR~\textcolor{trolleygrey}{(ours)} & \textbf{34.92} & \textbf{0.956} & \textbf{32.48} & \textbf{0.947} & \textbf{28.82} & \textbf{0.832}\\
\cmidrule(l){1-7}
Chung {\em et al.}\cite{chung2022improving} & 26.01 & 0.838 & 24.55 & 0.823 & 21.59 & 0.706 \\
Lahiri {\em et al.}~\cite{jin2017deep} & \underline{28.08} & \underline{0.931} & \underline{26.02} & \underline{0.856} & \underline{23.24} & \underline{0.812}\\
Zhang {\em et al.}~\cite{lahiri2022sparse} & 26.76 & 0.879 & 25.77 & 0.874 & 22.92 & 0.841\\
ADMM TV & 23.19 & 0.793 & 22.96 & 0.758 & 19.95 & 0.782\\
\bottomrule
\end{tabular}
}
\caption{
Quantitative evaluation of LA-CT (90$^\circ$) (PSNR, SSIM) on the AAPM 256$\times$256 test set. \textbf{Bold}: Best, \underline{under}: second best.
}
\vspace{-0.5cm}
\label{tab:la-ct}
\end{table}
The results of the limited angle tomography is presented in Table~\ref{tab:la-ct} and Fig.~\ref{fig:lact_results}. We test on the case where we have measurements in the $[0, 90]^{\circ}$ regime, and no measurements in the $[90, 180]^{\circ}$ regime. Hence, the task is to {\em infill} the missing views. Consistent with what was observed from SV-CT experiments, we see that DiffusionMBIR improves over the conventional diffusion model-based method~\cite{chung2022improving}, and also outperforms other fully supervised methods, where we see even larger gaps in performance between the proposed method and all the other methods. Notably, Chung et al.~\cite{chung2022improving} leverages no information from the adjacent slices, and hence has high degree of freedom on how to infill the missing angle. As the reconstruction is stochastic, we cannot impose consistency across the different slices. Often, this results in the structure of the torso being completely distorted, as can be seen in the first row of Fig.~\ref{fig:lact_results} (d). In contrast, our augmented prior imposes smoothness across frames, and also naturally robustly preserves the structure.

\noindent
\textbf{Compressed Sensing MRI.~}
\begin{table}[!t]
\centering
\setlength{\tabcolsep}{0.2em}
\resizebox{0.45\textwidth}{!}{% <------ Don't forget this %
\begin{tabular}{lllllll}
\toprule
{} & \multicolumn{2}{c}{\textbf{Axial$^*$}} & \multicolumn{2}{c}{\textbf{Coronal}} & \multicolumn{2}{c}{\textbf{Sagittal}} \\
\cmidrule(lr){2-3}
\cmidrule(lr){4-5}
\cmidrule(lr){6-7}
{\textbf{Method}} & {PSNR $\uparrow$} & {SSIM $\uparrow$} & {PSNR $\uparrow$} & {SSIM $\uparrow$} & {PSNR $\uparrow$} & {SSIM $\uparrow$}\\
\midrule
DiffusionMBIR~\textcolor{trolleygrey}{(ours)} & \textbf{41.49} & \textbf{0.974} & \textbf{37.36} & \textbf{0.942} & \textbf{37.18} & \textbf{0.953}\\
\cmidrule(l){1-7}
Score-MRI\cite{chung2022score} & \underline{40.38} & 0.968 & \underline{33.97} & \underline{0.925} & \underline{34.02} & \underline{0.928} \\
DuDoRNet~\cite{lahiri2022sparse} & 39.78 & \underline{0.974} & 33.56 & 0.927 & 33.48 & 0.927\\
Unet~\cite{jin2017deep} & 37.15 & 0.929 & 31.56 & 0.899 & 30.90 & 0.816\\
Zero-filled & 34.18 & 0.923 & 29.53 & 0.897 & 27.82 & 0.903\\
\bottomrule
\end{tabular}
}
\caption{
Quantitative evaluation of CS-MRI (acc. $\times$2) (PSNR, SSIM) on the BRATS data. \textbf{Bold}: Best, \underline{under}: second best.
}
\vspace{-0.5cm}
\label{tab:cs-mri}
\end{table}
We test our method on the reconstruction of 1D uniform random sub-sampled images, as was used in~\cite{zbontar2018fastmri}. Specifically, we keep 15\% of the autocalibrating signal (ACS) region in the center, and retain only the half of the k-space sampling lines, corresponding to approximately 2$\times$ acceleration factor for the acquisition scheme. The results are presented in Table~\ref{tab:cs-mri} and {supplementary} Fig.~\ref{fig:mri_results}. Consistent with what was observed in the experiments with SV-CT and LA-CT, we observe large improvements over the prior arts.

\noindent
\textbf{Out-of-distribution performance.~}
It was shown in the context of CS-MRI that diffusion models are surprisingly robust to the out-of-distribution (OOD) data~\cite{chung2022score,jalal2021robust}. For example, the score function trained with proton-density weighted, coronal knee scans only was able to generalize to nearly all the different anatomy and contrasts that were never seen at the test time. Would such generalization capacity also hold in the context of CT?  Here, we answer with a positive, and show that with the proposed method, one can use the same score function even when the targeted ground truth is vastly different from those in the training dataset. 

In Fig.~\ref{fig:ood}, we show two different cases (i.e. mild, severe) of such OOD reconstruction. For both cases, we see that 8-view is enough to produce high fidelity reconstructions that closely estimate the ground truth. Note that our prior is constructed from the anatomy of the human body, which is vastly different from what is given in Fig.~\ref{fig:ood} (b). Intuitively, in the Bayesian perspective, this means that our diffusion prior would have placed very little mass on images such as Fig.~\ref{fig:ood} (b). Nonetheless, we can interpret that diffusion models tend to place {\em some} mass even to these heavily OOD regions. Consequently, when incorporated with enough likelihood information, it is sufficient to guide proper posterior sampling, as seen in this experiment. Such property is particularly useful in medical imaging, where training data mostly consists of normal patient data, and the distribution is hence biased towards images without pathology. When at test time, we are given a patient scan that contains lesions, we desire a method that can fully generalize in such cases.

\noindent
\textbf{Choice of augmented prior.~}
%
% \begin{figure}[!b]
%     \centering
%     \includegraphics[width=0.5\textwidth]{./figs/ablation_TVdir.jpg}
%     \caption{Ablation study for the choice of augmented prior. (a) TV ($xyz$) prior, (b) TV ($z$) prior; proposed method, (c) Ground truth.}
%     \label{fig:ablation_TVdir}
% \end{figure}
%
In this work, we proposed to augment the diffusion generative prior with the model-based TV prior, in which we chose to impose the TV constraint only in the redundant $z-$direction, while leaving the $xy-$plane intact. This design choice stems from our assumption that diffusion prior {\em better} matches the actual prior distribution than the TV prior, and mixing with the TV prior might compromise the ability of diffusion prior.

To verify that this is indeed the case, we conducted an ablation study on TV($xyz$) and TV($z$) prior. {Supplementary} Fig.~\ref{fig:ablation_TVdir} shows the visual and numerical results on the priors. Both priors have scored high on PSNR and SSIM, but we can figure out that the images with TV prior on $xy$-plane are blurry compared to the samples from the proposed one. The analysis implies that the usage of TV($xyz$) prior shifted the result a bit far apart from the well-trained diffusion prior.
\vspace{-0.3cm}

\section{Conclusion}
\label{sec:conclusion}
\vspace{-0.2cm}

In this work, we propose DiffusionMBIR, a diffusion model-based reconstruction strategy for performing 3D medical image reconstruction. We show that all we need is a 2D diffusion model that can be trained with little data ($< 10$ volumes), augmented with a classic TV prior that operates on the redundant $z$ direction. We devise a way to seamlessly integrate the usual diffusion model sampling steps with the ADMM iterations in an efficient way. The results demonstrate that the proposed method is capable of achieving state-of-the-art reconstructions on  Sparse-view CT, Limited-angle CT, and Compressed-sensing MRI. Specifically for Sparse-view CT, we show that our method is capable of providing accurate reconstructions even with as few as two views. Finally, we show that DiffusionMBIR is capable of reconstructing OOD data that is vastly different from what is presented in the training data.

%-------------------------------------------------------------------------

%%%%%%%%% REFERENCES
{\small
\bibliographystyle{ieee_fullname}
\bibliography{egbib}
}

\clearpage
%%%%%%%%%%%%%%%%%%
% Supplementary material starts here
%%%%%%%%%%%%%%%%%%
\appendix

\section*{\Large{\textbf{Supplementary Material}}}

\section{ADMM-TV}
\label{sec:admm_tv}

In this section, we derive the ADMM-TV optimization framework for completeness.
We are interested in solving the problem of TV-regularized WLS of the following form:
\begin{align}
    \min_\xb \frac{1}{2}\|\yb - \Ab\xb\|_2^2 + \lambda \|\Db_z\xb\|_1,
\end{align}
where $\Db_z$ takes the finite difference across the $z-$dimension. In order to solve the problem in an alternating fashion, we split the variables
\begin{align}
    \min_{\xb,\zb} \quad & \frac{1}{2}\|\yb - \Ab\xb\|_2^2 + \lambda\|\zb\|_1 \\
    \textrm{s.t.} \quad & \zb = \Db_z\xb.
\end{align}
The scaled formulation of ADMM~\cite{boyd2011distributed} is then given by
\begin{align}
    \xb^+ &= \argmin_{\xb} \frac{1}{2}\|\yb - \Ab\xb\|_2^2 + \frac{\rho}{2}\|\Db_z\xb - \zb + \wb\|_2^2 \label{eq:gen_admm_x}\\
    \zb^+ &= \argmin_{\zb} \lambda\|\zb\|_1 + \frac{\rho}{2}\|\Db_z\xb^+ - \zb + \wb\|_2^2 \label{eq:gen_admm_z}\\
    \wb^+ &= \wb + \Db_z\xb^+ - \zb^+. \label{eq:gen_admm_w}
\end{align}
\eqref{eq:gen_admm_x} is convex and smooth, and thus has a closed form solution
\begin{align}
    \xb^+ = (\Ab^T\Ab + \rho \Db_z^T\Db_z)^{-1}(\Ab^T \yb + \rho \Db_z^T(\zb - \wb)),
\end{align}
where one can perform CG rather than computing the matrix inverse directly.
In order to solve, \eqref{eq:gen_admm_z}, we define the proximal operator~\cite{parikh2014proximal} as
\begin{align}
    {\rm{prox}}_{f, \eta}(\zb) \triangleq \argmin_{\xb} f(\xb) + \frac{1}{2\eta} \|\xb - \zb\|_2^2.
\end{align}
By inspecting \eqref{eq:gen_admm_z}, we know that it is in the form of proximal mapping
\begin{align}
    \zb^+ &= {\rm{prox}}_{\|\cdot\|_1, \lambda/\rho}(\Db_z\xb + \wb)\\
    &= \Sc_{\lambda/\rho}(\Db_z\xb + \wb),
\end{align}
where we have leveraged the fact that the proximal mapping of the $\ell_1$ norm is given as the soft thresholding operator $\Sc$. In summary, we have
\begin{align*}
    \xb^{+} &= (\Ab^T\Ab + \rho \Db_z^T\Db_z)^{-1}(\Ab^T\yb + \rho \Db_z^T(\zb - \wb))\\
    \zb^{+} &= \Sc_{\lambda/\rho}(\Db_z \xb^{+} + \wb)\\
    \wb^{+} &= \wb + \Db_z \xb^{+} - \zb^{+}.
\end{align*}
The algorithmic detail can be found in Algorithm~\ref{alg:diffusion_mbir_slow}.

\begin{algorithm}[!hbt]
\caption{DiffusionMBIR (slow)}
\begin{algorithmic}[1]
\Require $s_{\theta}, N, M, K, \lambda, \rho, \{\sigma_i\}$
\State $\xb_N \sim \Nc({\bf 0}, \sigma_{T}^{2} \Ib)$
\For{$i = N-1:0$} \Comment{SDE iteration} \do \\
\State {$\bar\xb_i \gets {\rm Solve}(\xb_{i+1}, \s_{\theta^*})$}
\State $\Ab_{\rm{CG}} \gets \Ab^T\Ab + \rho \Db_z^T\Db_z$
\State $\zb^{(1)} \gets$ \code{torch.zeros\_like}$(\bar\xb_i)$
\State $\wb^{(1)} \gets$ \code{torch.zeros\_like}$(\bar\xb_i)$
\For{$j = 1:M$} \do \\ \Comment{ADMM iteration}
\State $\bb_{\code{CG}}^{(j)} \gets \Ab^T\yb + \rho \Db^T(\zb^{(j)} - \wb^{(j)})$
\State $\bar\xb_i^{(j+1)} \gets {\code{CG}}(\Ab_{\code{CG}}, \bb_{\code{CG}}^{(j)}, K)$ \Comment{CG iteration}
\State $\zb^{(j+1)} \gets \Sc_{\lambda/\rho}(\Db_z \bar\xb_i^{(j+1)} + \wb^{(j)})$
\State $\wb^{(j+1)} \gets \wb^{(j)} + \Db_z \bar\xb_i^{(j+1)} - \zb^{(j+1)}$
\EndFor
\State $\xb_i \gets \bar\xb_i^{(M+1)}$
\EndFor
\State \textbf{return} $\xb_0$
\end{algorithmic}\label{alg:diffusion_mbir_slow}
\end{algorithm}

\section{Details of experiment}
\label{sec:supp_exp_details}

\begin{figure*}[!hbt]
    \centering
    \includegraphics[width=1.0\textwidth]{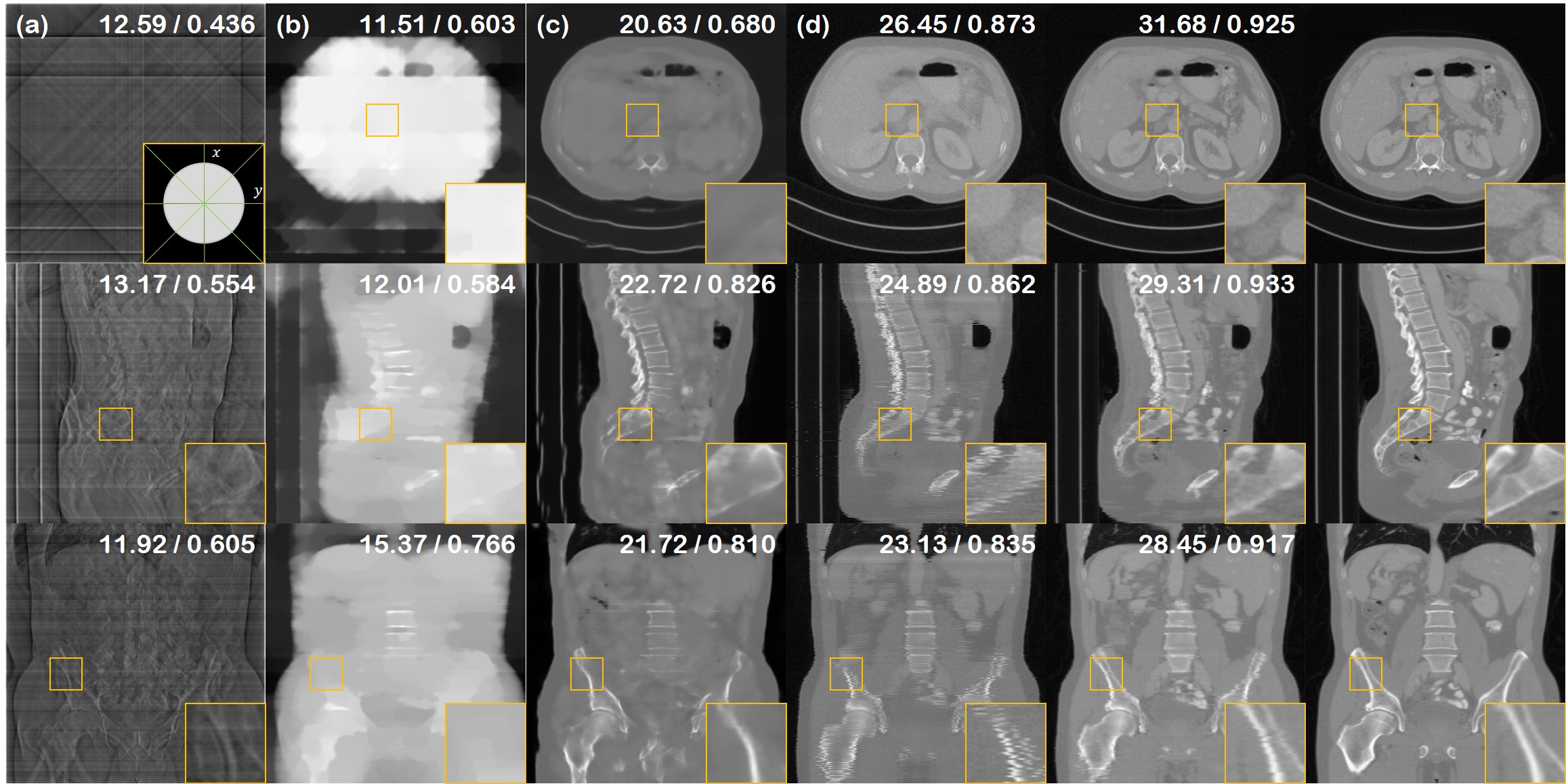}
    \caption{4-view SV-CT reconstruction results of the test data (First row: axial slice, second row: sagittal slice, third row: coronal slice). (a) FBP, (b) ADMM-TV, (c) Lahiri {\em et al.}~\cite{lahiri2022sparse}, (d) Chung {\em et al.}~\cite{chung2022improving}, (e) proposed method, (f) ground truth. Green lines in the inset of first row (a): measured angles.}
    \label{fig:4view_results}
\end{figure*}

\subsection{Dataset}

\noindent
\textbf{AAPM.~}
\begin{table}[]
\centering
\setlength{\tabcolsep}{8pt}
\resizebox{0.48\textwidth}{!}{% <------ Don't forget this %
\begin{tabular}{lccccc}
\toprule
Patient ID & \# slices & \thead{FOV \\ (mm)}& KVP & \thead{Exposure\\time (ms)} & \thead{X-ray \\tube current (mA)}\\
\midrule
L067 & 310 & 370 & 100 & 500 & 234.1\\
L109 & 254 & 400 & 100 & 500 & 322.3\\
L143 & 418 & 440 & 120 & 500 & 416.9\\
L192 & 370 & 380 & 100 & 500 & 431.6\\
L286 & 300 & 380 & 120 & 500 & 328.9\\
L291 & 450 & 380 & 120 & 500 & 322.7\\
L310 & 340 & 380 & 120 & 500 & 300.0\\
L333 & 400 & 400 & 100 & 500 & 348.7\\
L506 & 300 & 380 & 100 & 500 & 277.7\\
\midrule
L097 (test) & 500 & 430 & 120 & 500 & 327.6\\
\bottomrule
\end{tabular}
}
\vspace{0.2em}
\caption{
AAPM dataset specification. L097 volume is used for testing while the other volumes are used for training. 
}
\label{tab:dataset}
\end{table}
\begin{figure}[!hbt]
    \centering
    \includegraphics[width=0.45\textwidth]{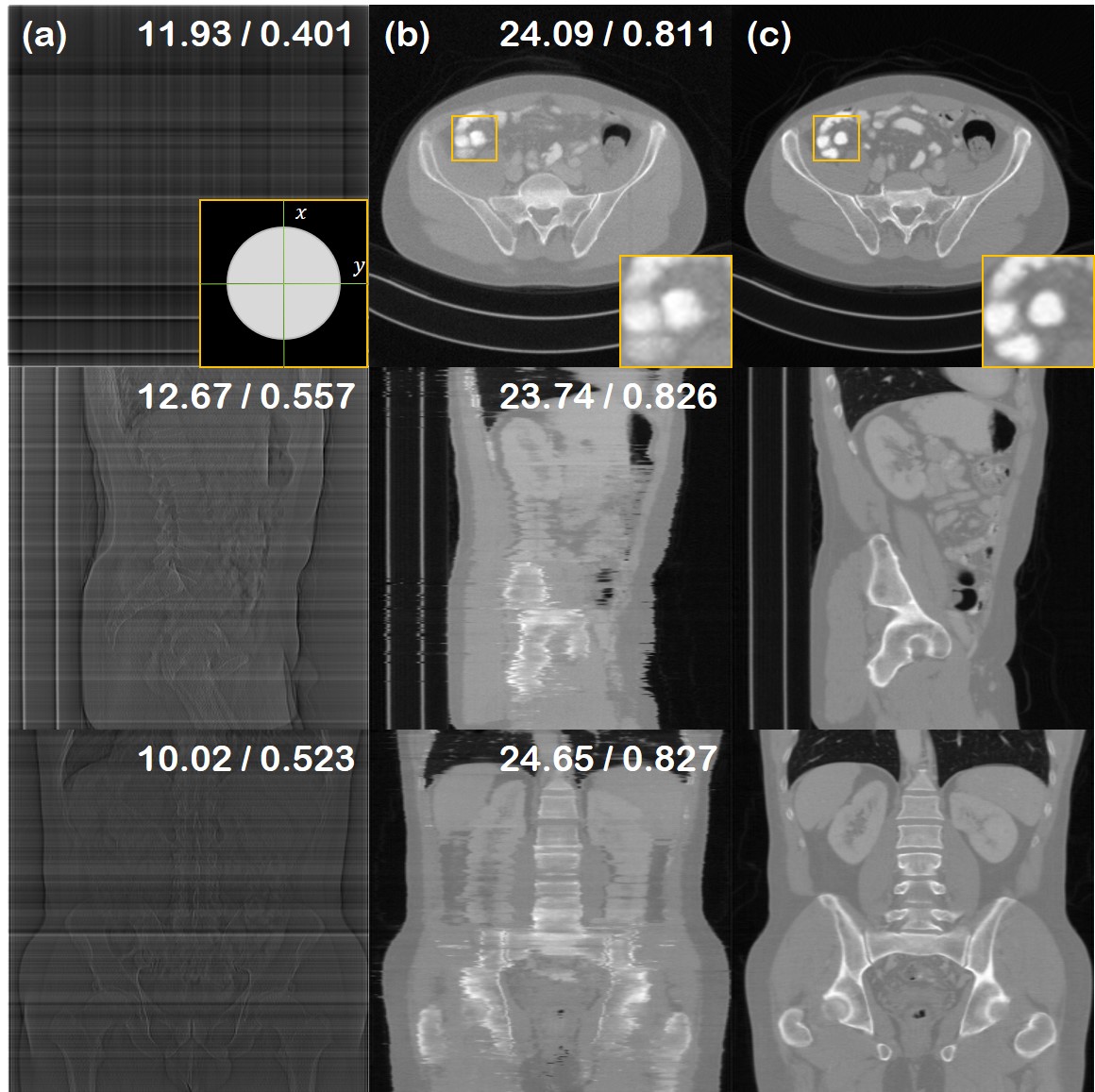}
    \caption{2-view SV-CT reconstruction results of the test data (First row: axial slice, second row: sagittal slice, third row: coronal slice). (a) FBP, (b) proposed method, (f) ground truth. Green lines in the inset of first row (a): measured angles.}
    \label{fig:2view_results}
\end{figure}
We take the dataset from the AAPM 2016 CT low-dose grand challenge, where the data are acquired in a fan-beam geometry with varying parameters, as presented in Table~\ref{tab:dataset}. The data preparation steps follow that of~\cite{kang2017deep}. From the helical cone beam projections, approximation to fanbeam geometry is performed via single-slice rebinning technique~\cite{noo1999single}. Reconstruction is then performed via standard filtered backprojection (FBP), where the reconstructed axial images have the matrix size of $512\times512$. We resize the axial slices to have the size $256\times256$, and use these slices to train the score function. The whole dataset consists of 9 volumes (3142 slices) of training data, and 1 volume (500 slices) of testing data. To generate sparse-view measurements, we retrospectively employ the parallel-view geometry for simplicity.

\noindent
\textbf{BRATS.~}
We take the dataset from the multimodal brain tumor segmentation BRATS 2018 challenge ~\cite{menze2014multimodal}, where we select the test data as the first FLAIR volume, which has the matrix size of $240\times240\times154$. As stated in the main text, all the methods were trained with the separate fastMRI 2019 knee database~\cite{zbontar2018fastmri}.

\subsection{Details of network training}

For the CT score function, we train the \code{ncsnpp} network~\cite{song2020score} without modifications with \eqref{eq:dsm} by setting $\lambda = \sigma^2(t)$~\cite{song2021maximum}, and $\varepsilon = 10^{-5}$. Our network is trained using the Adam optimizer ($\beta_1 = 0.9, \beta_2 = 0.999$) with a linear warm-up schedule, reaching $2\times10^{-4}$ at the 5000$^{\rm{th}}$ step, and with a batch size of 2 using a single RTX 3090 GPU for 200 epochs. Training took about a week and a half. 
%We will provide a reproducible training script in our open-sourced code repository.

\subsection{Comparison methods}

\noindent
\textbf{Chung {\em et al.}~\cite{chung2022improving}}
As the method is based on diffusion models, we use the same pre-trained score function, and use the reconstruction scheme of~\cite{chung2022improving}, which amounts to applying ART after every PC update steps.

\noindent
\textbf{Lahiri {\em et al.}}~\cite{lahiri2022sparse}
We use 2-stage reconstructing 3D CNNs, where we train the networks with slabs, not taking patches in the xy dimension, as the paper suggests. The architecture of CNN was taken as a standard U-Net~\cite{zbontar2018fastmri} architecture rather than stack of single-resolution CNNs, as we achieved better performance with U-Nets. Furthermore, we drop the adversarial loss and only use the standard reconstruction loss, as we found the training to be more stable. CG was applied with 30 iterations.

\noindent
\textbf{FBPConvNet~\cite{jin2017deep}, Zhang {\em et al.}}~\cite{zhang2016image}
We use the same U-Net architecture that was used to train Lahiri {\em et al.}~\cite{lahiri2022sparse}, but only on 2D images. Note that the original work of Zhang {\em et al.}~\cite{zhang2016image} uses a much simpler CNN architecture, which leads to degraded performance.

\noindent
\textbf{ADMM-TV.~}
We minimize the following objective
\begin{align}
    \min_\xb \frac{1}{2}\|\yb - \Ab\xb\|_2^2 + \lambda \|\Db\xb\|_{2, 1},
\end{align}
where $\Db = [\Db_x, \Db_y, \Db_z]$, which corresponds to the isotropic TV. The outer iterations are solved with ADMM (30 iterations), while the inner iterations are solved with CG (20 iterations). We perform coarse grid search to find the parameter values that produce low MSE values, then perform another grid search to find the most visually pleasing solution --- images with salient edges.
The parameter is set to $(\lambda, \rho) = (0.5, 50)$ for SV-CT, and $(\lambda, \rho) = (0.15, 40)$ for LA-CT.

\noindent
\textbf{Score-MRI.~}\cite{chung2022score}
The method utilizes the same score function as the proposed method, but relies on iterated projections onto the measurement subspace after every iteration of the PC update. Every slices in the xy dimension are reconstructed separately, then stacked to form the volume.

\noindent
\textbf{DuDoRNet.~}\cite{zhou2020dudornet}
We train DuDoRNet with 4 recurrent blocks and the default parameters, following the configuration of~\cite{chung2022score}. The network is trained on the fastMRI knee dataset, with the proton density (PD) / proton density fat suppressed (PDFS) image for the prior information.

\noindent
\textbf{U-Net.~}\cite{zbontar2018fastmri}
We leverage the pre-trained U-Net on the fastMRI dataset, trained only with the L1 loss in the image domain.

\begin{figure*}[!t]
    \centering
    \includegraphics[width=1.0\textwidth]{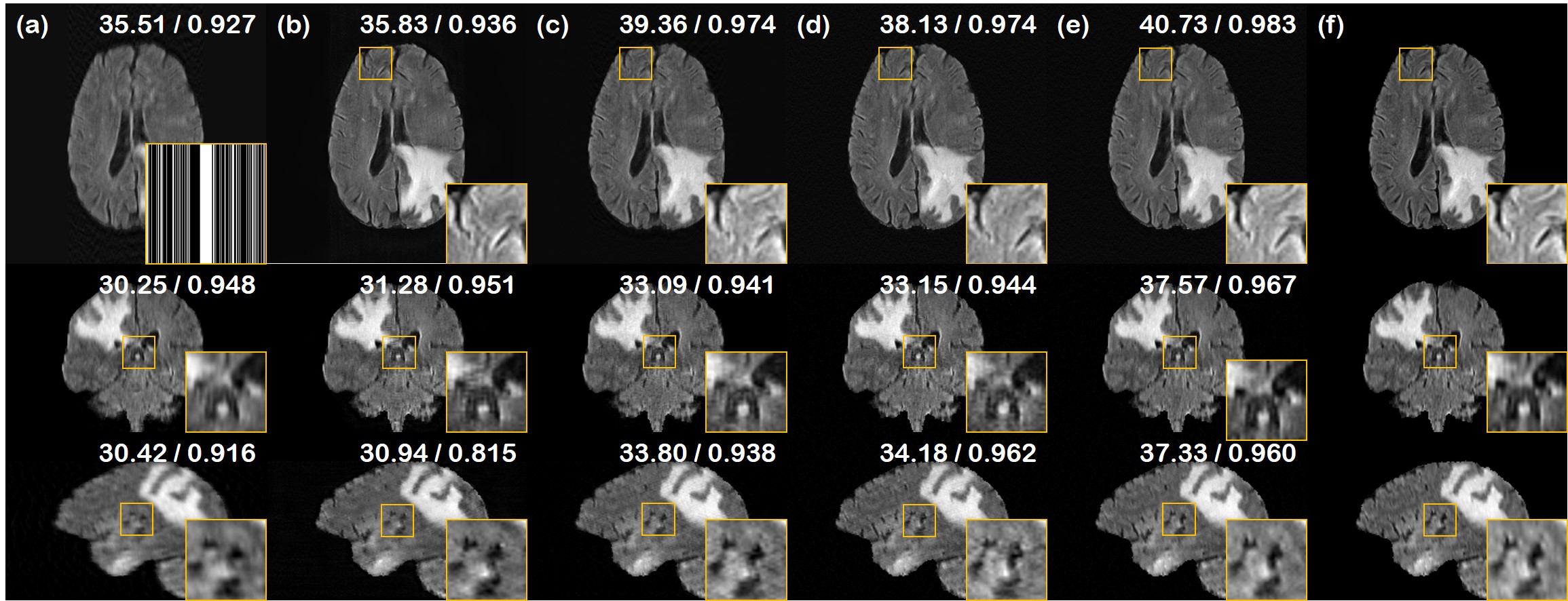}
    \caption{CS-MRI results of the test data (First row: axial slice, second row: sagittal slice, third row: coronal slice). (a) zero-filled, (b) U-Net~\cite{zbontar2018fastmri}, (c) DuDoRNet~\cite{zhou2020dudornet}, (d) Score-MRI~\cite{chung2022score}, (e) proposed method, (f) ground truth. PSNR/SSIM values presented in the upper right corner. Mask presented in the first row of (a): Sub-sampling mask applied to all slices.}
    \label{fig:mri_results}
\end{figure*}

\section{Further experiments}
\label{sec:supp_experiments}

\subsection{Additional experimental results}

4-view sparse view tomographic reconstruction is presented in Fig.~\ref{fig:4view_results}. Furthermore, we demonstrate that we can even perform {\em 2-view} reconstruction, as can be seen in Fig.~\ref{fig:2view_results}. In this regime, the information contained in the measurement is very few and sparse --- clearly not sufficient for achieving an {\em accurate} reconstruction. As we are leveraging the generative prior however, we can sample multiple reconstructions that are 1) perfectly measurement feasible, and 2) looks realistic. Although this might not be of significant importance in the medical imaging field, it could greatly impact fields where approximate reconstructions from very limited acquisitions are necessary.

\subsection{Number of views vs. performance}
\label{sec:nview_vs_psnr}

\begin{figure}\adjustbox{width=0.9\columnwidth}{
\centering
\begin{tikzpicture}
	\begin{axis}[
	height=5cm, width=6cm,font=\tiny,
		xmin={1.7}, xmax={40}, xmode={log}, xtick={2, 4, 8, 16, 32}, xticklabels={\tickNview{2}, $4$, $8$, $16$, $32$},
		ymin={20.0}, ymax={38.0}, ytick={22.0, 26.0, 30.0, 34.0, 37.0}, yticklabels={$22.0$, $26.0$, $30.0$, $34.0$, \tickPSNR},
  grid={major}, legend style={font=\tiny, at={(0.95, 0.6)}},
	]
	\addplot[C0,mark=o,mark size=1.5pt] coordinates {
		(2, 24.42)
		(4, 30.25)
		(8, 33.16)
		(16, 34.18)
		(32, 34.52)
	};
	\addlegendentry{Axial}
	\addplot[C1,mark=o,mark size=1.5pt,mark=square] coordinates {
		(2, 23.16)
		(4, 30.09)
		(8, 35.17)
		(16, 36.84)
		(32, 37.01)
	};
	\addlegendentry{Coronal}
	\addplot[C2,mark=o,mark size=1.5pt,mark=*] coordinates {
		(2, 21.72)
		(4, 28.94)
		(8, 32.16)
		(16, 33.77)
		(32, 34.19)
	};
	\addlegendentry{Sagittal}
	\end{axis}
\end{tikzpicture}
}
\captionof{figure}{Number of measured views vs. PSNR[db]}
\label{fig:nview_vs_psnr}
\end{figure}
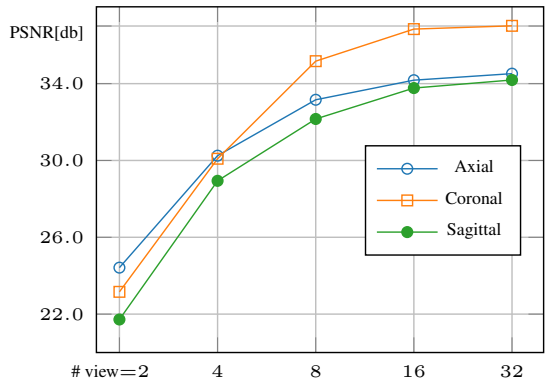

We have verified that we can perform extreme sparse-view reconstruction with the proposed method. One can naturally ask the limit of the proposed method, and the
trend between the measured number of views versus the reconstruction performance, which we show in the plot shown in Fig.~\ref{fig:nview_vs_psnr}. Note that the given plot is a {\em log} plot in the $x$-axis (i.e. \# views). We can easily see that the performance caps if we increase the number of measurements to higher than 16, and we also see that down to 8-views, we can acquire reconstructions with only a modest drop in the performance. The performance starts to heavily degrade as we drop down the number of views below 4. We conclude that there is a singular point, where the information in the measurement is just not enough, even when we have a very strong reconstruction algorithm.
\begin{figure}[!b]
    \centering
    \includegraphics[width=0.5\textwidth]{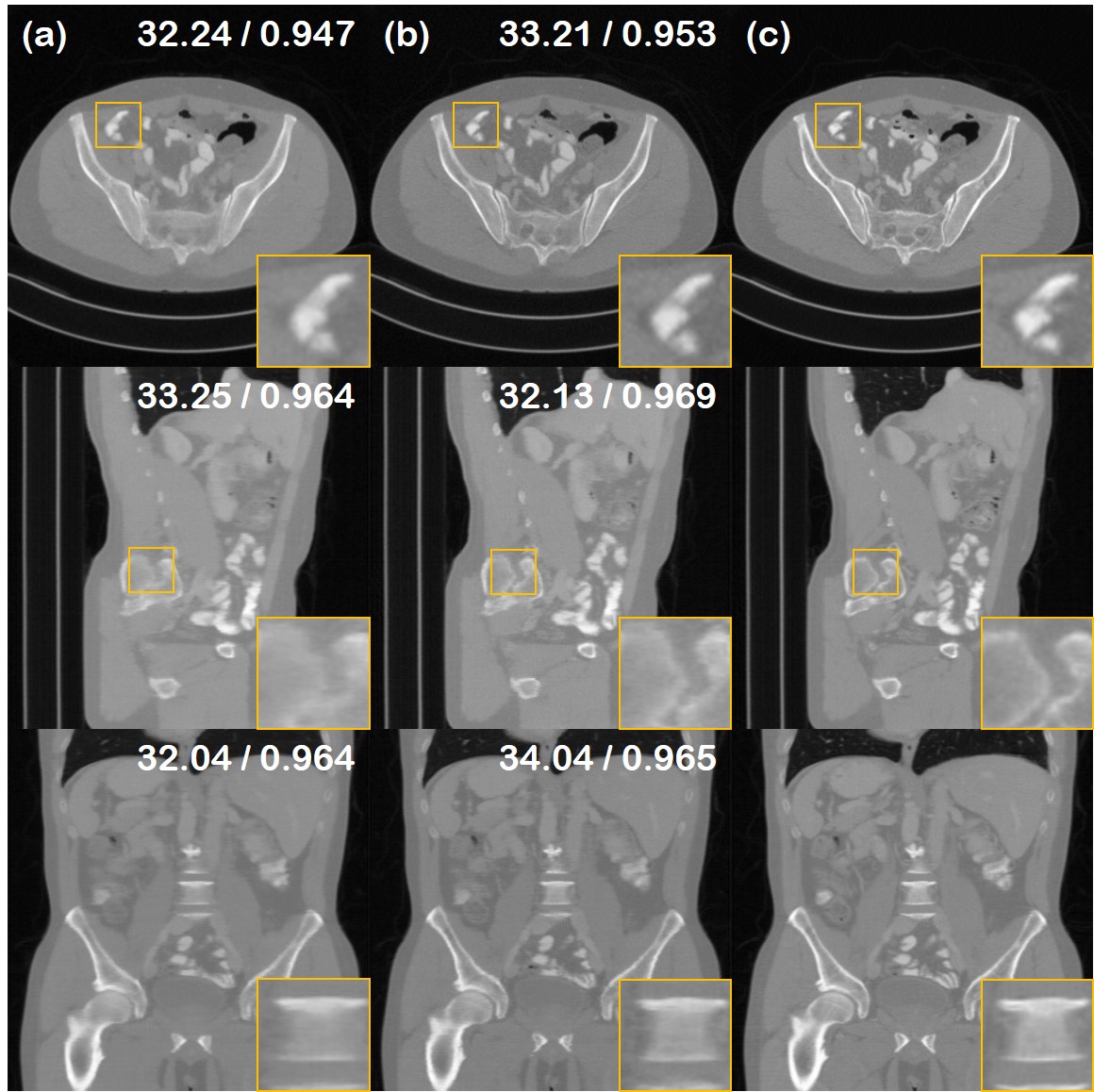}
    \caption{Ablation study for the choice of augmented prior. (a) TV ($xyz$) prior, (b) TV ($z$) prior; proposed method, (c) Ground truth.}
    \label{fig:ablation_TVdir}
\end{figure}

\end{document}